\documentclass{spie}

\usepackage{cite}

\usepackage{times} 
\usepackage{indentfirst} 

\usepackage{wrapfig}
\usepackage{siunitx}
\usepackage{amsfonts}
\usepackage{amsmath}
\usepackage{enumitem}
\usepackage{booktabs} 
\usepackage{multirow}
\usepackage{hyperref}

\DeclareMathOperator*{\argmin}{arg\,min}

\usepackage[colorinlistoftodos,color=pink]{todonotes} 
\usepackage{xkeyval}
\presetkeys{todonotes}{inline}{}

\usepackage{lipsum} 

\title{Wind Turbine Feature Detection Using Deep Learning and Synthetic Data}

\author{Arash Shahirpour\supit{1}, Jakob Gebler\supit{1}, Manuel Sanders\supit{1}, Tim Reuscher\supit{1}
  \skiplinehalf
  \normalsize 
  \supit{1} Institute of Automatic Control (IRT), RWTH Aachen University
}

\begin{document}

\maketitle

\begin{abstract}
For the autonomous drone-based inspection of wind turbine (WT) blades, accurate detection of the WT and its key features is essential for safe drone positioning and collision avoidance. Existing deep learning methods typically rely on manually labeled real-world images, which limits both the quantity and the diversity of training datasets in terms of weather conditions, lighting, turbine types, and image complexity. In this paper, we propose a method to generate synthetic training data that allows controlled variation of visual and environmental factors, increasing the diversity and hence creating challenging learning scenarios. Furthermore, we train a YOLOv11 \cite{jocher_ultralytics_2024} feature detection network solely on synthetic WT images with a modified loss function, to detect WTs and their key features within an image. The resulting network is evaluated both using synthetic images and a set of real-world WT images and shows promising performance across both synthetic and real-world data, achieving a Pose mAP50-95 of 0.97 on real images never seen during training. 
\end{abstract}

\section{Introduction}
\begin{wrapfigure}[19]{r}{0.55\textwidth}
  \vspace{-10pt} 
  \centering
  \includegraphics[width=0.53\textwidth]{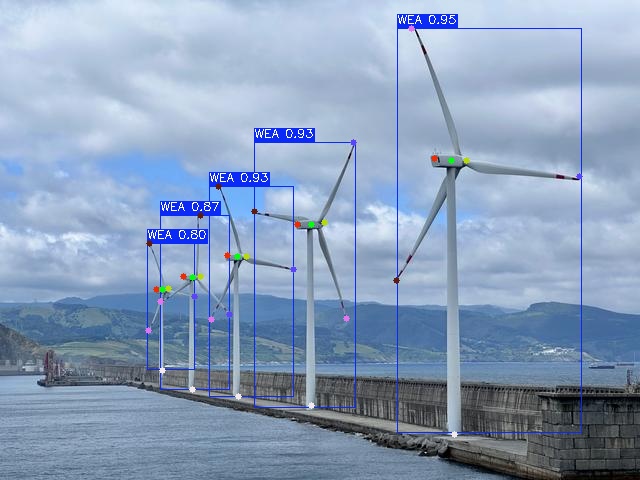}
  \caption{Predicted wind turbine keypoints on unseen real-world data. Raw image from~\cite{unsplash_wind_nodate} with Unsplash license.}
  \label{fig:5:nices_Bild}
  \vspace{20pt} 
\end{wrapfigure}
With the increasing global demand for renewable energy, wind turbines (WTs) have become the fastest-growing renewable energy technology over the past four decades \cite{enevoldsen_examining_2019}. As the number of WTs continues to increase worldwide, maintenance has become one of the primary challenges facing this industry. Since WTs are subject to harsh weather conditions and events such as bird collisions and lightning strikes, WTs are prone to surface and structural damage \cite{moolan-feroze_simultaneous_2019}. Consequently, reducing maintenance duration and costs has become the main objective of current maintenance strategies \cite{tchakoua_wind_2014}. Fig.~\ref{fig:5:nices_Bild} shows such WTs. Regular inspections have the potential to play a critical role in the early detection of surface damage and in minimizing WT downtime \cite{garcia_marquez_condition_2012}. However, current inspection methods primarily rely on manual techniques such as climbing equipment, manually operated drones, or ground-based photos \cite{moolan-feroze_simultaneous_2019}. As these methods are costly, time-consuming, and pose safety issues, there is a growing interest in automating the WT surface inspection process \cite{shihavuddin_wind_2019}. 

One increasingly popular method, due to its reduced costs and increased safety compared to manual techniques, is to employ automated drones for the inspection. In addition to capturing WT surface data to perform the inspection, these drones must navigate safely and position themselves in front of the relevant WT components accurately while avoiding collision. Although drones are typically equipped with Global Navigation Satellite Systems (GNSS) antennas and the positions of the WTs in the absence of wind are known in advance, the real-time position of the WT towers, which is deflected by wind, and more importantly, their blades, is generally not available.

The goal in this study is to use RGB (red, green, blue) camera images to extract key features of operational WTs that can be used to estimate the position and yaw angle of WTs relative to the drone. A preview of our keypoint detection results is shown in Fig.~\ref{fig:5:nices_Bild}, illustrating our method's ability to localize multiple wind turbines and their features. This research is part of the AutoFlow project (this project is supported by the Federal Ministry for Economic Affairs and Energy of Germany (BMWE) with the project number 03EE3064F). 

\begin{wrapfigure}[31]{r}{0.3\textwidth}
  \vspace{-10pt} 
  \centering
  \includegraphics[width=0.2\textwidth]{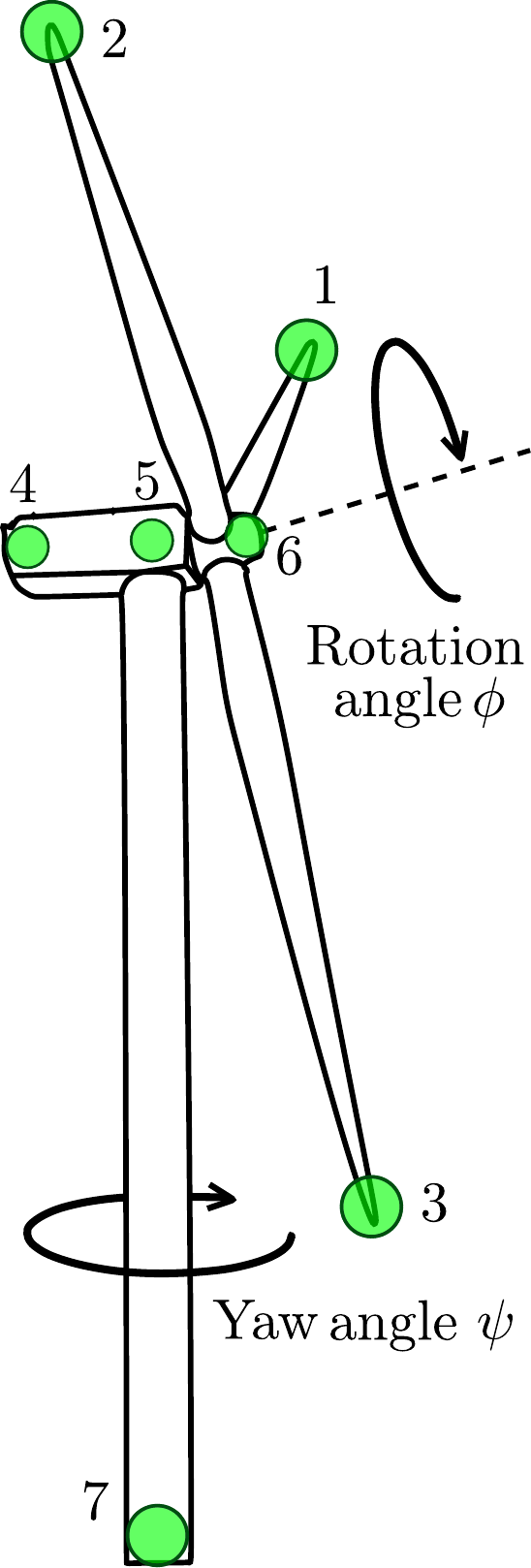}
  \caption{A wind turbine with the possible rotations during the inspection: yaw angle $\psi$ and rotation angle of the blades $\phi$. The key features of a wind turbine are marked by green circles and include the tips of the blades, the front and rear ends of the hub, and the top and bottom ends of the tower.}
  \label{fig:01_WEA}
  \vspace{20pt} 
\end{wrapfigure}

 Since the WT is operational during inspection, its yaw angle may change due to a change in wind direction. Fig.~\ref{fig:01_WEA} illustrates the possible WT rotations and highlights the seven visual features of a WT that must be extracted from each image. In the following, the state of the art in this research field is presented and discussed.  

In works such as \cite{stokkeland_computer_2014}, \cite{stokkeland_autonomous_2015}, and \cite{rao_wind_2019}, traditional computer vision methods are employed to first detect features of a WT, such as the long blade and tower lines, to eventually find the WT tower. While these methods deliver promising results, there are limitations to their use cases. These methods function best under clear weather conditions where the blue sky serves as a good contrast to the wind turbine. Cloudy weather and different light conditions might cause challenges. This is a known challenge with traditional computer vision methods, as they are not reliable for all conditions and require more fine-tuning for different conditions \cite{omahony_deep_2020}. As a result, the majority of current approaches rely on deep learning instead.

In \cite{moolan-feroze_simultaneous_2019}, a least squares optimization method is combined with a CNN to infer projections of a skeletal model of a WT to the camera images. The objective is to improve the drone's localization around the WT. One drawback of this work is that the real-world testing is limited to a single WT, which restricts the real-world generalizability and applicability of this method.

An attempt to detect the tips of WT blades is presented in \cite{guo_detecting_2019}, where a CNN is combined with a FAST corner detector to detect the pixels of the tips. By solving the resulting Perspective-n-Point (PnP) problem, the 3D world coordinates of the tips are estimated. Another method is presented in \cite{gu_autonomous_2020}, where the You Only Look Once (YOLO) v3 is used to detect the WT and its hub, followed by a Hough transform algorithm to detect the blades. However, both these works rely on manually labeled images and focus on non-operational WTs where the rotors stand still during inspection.

The state-of-the-art approaches present several limitations. First, since the images are collected mostly in one WT farm using a single camera, they are prone to being limited in diversity in terms of weather and lighting conditions, WT types, and background complexity. These can negatively impact the generalization of the trained network. Second, incorporating these features into a dataset may require scaling up the number of training images. However, manual annotation is time-consuming and does not scale well to larger datasets, for example, those with 10,000 images. While this is a general challenge when training a network to detect arbitrary objects, the structured and largely uniform geometry of WTs, along with their predictable appearance, suggests that it may be feasible to train the network using synthetically rendered images rather than relying exclusively on real-world photos. Synthetic data also makes it easier to systematically control and vary visual features such as lighting, background complexity, and WT configurations.

The idea of training a network using synthetic data has been implemented in works such as \cite{tremblay_deep_2018}, \cite{tremblay_training_2018}, where a toolbox is developed for 3D pose estimation of known objects that uses a network trained exclusively on synthetic data. Another successful application of this approach is demonstrated in~\cite{hebisch_camera-based_2022}, where the authors trained a network to estimate the pose of a small aircraft flying above a camera mounted on a moving car. The network was trained exclusively on synthetic data. These approaches focus on fixed and well-defined objects with known geometry and dimensions, making synthetic training more straightforward within their specific scope.

Given the limitations of the state-of-the-art approaches, we introduce a flexible and scalable synthetic data generation toolbox. This toolbox makes it possible to render an arbitrary number of annotated images, covering a wide range of visual conditions, including diverse backgrounds, lighting effects, and multiple WTs per image with overlapping bounding boxes. This method is presented in Sec.~\ref{Data generation}. Using only the synthetic dataset, we train a YOLOv11-based keypoint detection network with a modified loss function. The network is capable of detecting the seven key features for multiple WTs within the same image. The training process is explained in detail in Sec.~\ref{training}. In addition to evaluating the model on the synthetically generated test set, we also assess generalization using a second dataset composed of real-world images collected from online sources. The experiments and validation are presented in Sec.~\ref{results}. Finally, the conclusion and outlook of our work are discussed in Sec.~\ref{conclusion}.

The primary contributions of this paper include a fully customizable rendering toolbox capable of generating inexpensive synthetic datasets along with corresponding annotation files, a YOLOv11 network trained exclusively on these synthetic images using a modified loss function to detect key points of wind turbines within an image, and the public release of both the toolbox and trained models on GitHub~\cite{shahirpour_wind_2025} to support reproducibility and further development.

\section{METHOD}
\subsection{Synthetic Data Generation} \label{Data generation}
To create the images, BlenderProc2~\cite{denninger_blenderproc2_2023} was used, which can create photorealistic renders of a scenery via the Python interface of the free and open-source 3D creation software Blender~\cite{blender_online_community_blender_nodate}. The Blender interface also allows to retrieve the coordinates of the elements in the scene and thus automatic label creation.

The project's initial aim was to develop an algorithm designed for a specific offshore WT farm. Consequently, the synthetic image generation toolbox was designed to replicate the visual characteristics and challenges of that particular site. While the offshore-specific functionality remains in the toolbox, the study strategy shifted towards increasing generality both in the WT models and in terms of the background. The goal was to ensure that the resulting network was not limited to a specific wind farm, WT size, and background. For more diverse WT models, the original set of the WT computer-aided design (CAD) models was expanded by adding additional variations. These variations differ in blade thickness and blade length.

To increase the generality of the background, images of WT farms with a transparent background were generated by BlenderProc2 while maintaining all other visual elements. These images were then combined with randomly selected landscape images as wind turbines are typically located in natural environments, and such landscape images reflect their typical visual context. For this goal, a collection of 4319 random background images was obtained from kaggle~\cite{kaggle_landscape_nodate} under CC0 public domain licensing. The resulting images contain realistic WTs placed against a wide variety of background scenes.

In addition to the WT sizes and backgrounds, variations of several key factors were considered in the final images to further increase generalization. These elements were chosen based on our field observations in different wind farms and discussions with wind farm operators. These parameters can be configured in the toolbox:

\begin{itemize}[noitemsep, topsep=0pt]
    \item \textbf{Sun position and lighting}: the sun's position and light intensity can be configured with solar azimuth $\phi_\text{s}$, altitude $\theta_\text{s}$, and dust density $d_\text{d}$. 
    \item \textbf{WTs number and positioning}: this is configured with the number $n$ and the position $(x,y)$ of the WTs in each image.
    \item \textbf{WT angles}: parameters yaw $\psi$ and blade rotation angle $\phi$ define the angles of the appearing WTs.
    \item \textbf{Camera}: the camera height from the ground $h_\text{c}$, camera distance from the origin of the Blender world $d_\text{c}$, and the focal length $f_\text{c}$ can be configured. A further parameter $c_\text{c}$ defines whether the camera is centered on the WT hub. Optionally, the camera can be configured with roll, pitch, and yaw angles $\phi_\text{c},\theta_\text{c},\psi_\text{c}$ for each image.
\end{itemize}
In addition to the image elements discussed above, the following augmentations were applied to the rendered images to further increase the variety in the resulting dataset: 
\begin{itemize}[noitemsep, topsep=0pt]
    \item \textbf{Hue Saturation Value (HSV) shifts:} small random shifts in the HSV values of the foreground and background image. The shift is the same for each pixel in an image and is configured by $H_\text{f},S_\text{f},V_\text{f}$ for the foreground and with $H_\text{b},S_\text{b},V_\text{b}$ for the background of the image. 
    \item \textbf{JPEG compression artifacts}: JPEG is a commonly used image format. JPEG compression $c_\text{JPEG}$ can be applied to represent these artifacts in the dataset.
    \item \textbf{Random noise}: random Gaussian distributed noise can be added to each pixel individually. The distribution is parameterized by mean $\mu_\text{n}$ and standard deviation $\sigma_\text{n}$.  
    \item \textbf{Random noise background}: the percentage of images created with a random noise image, as depicted in Fig.~\ref{fig:6_Validationset1} with the lower middle image, is configured by the parameter $n_\text{r}$.
\end{itemize}
 These parameters are summarized in Tab. 1. Given the specifications and features described above, the toolbox generates the desired number of images along with the corresponding annotation files. When generating images with the toolbox, the parameters can be varied for each image to create a diverse dataset. 
 Fig.~\ref{fig:6_Validationset1} shows a collage of the synthetic images generated by the toolbox and also inference results, which will be discussed separately in Sec.\ref{results}.

\subsection{Training a Pre-trained Convolutional Deep Neural Network}\label{training}
\begin{wrapfigure}[22]{r}{0.5\textwidth}
  \vspace{-10pt} 
  \centering
  \includegraphics[width=0.48\textwidth]{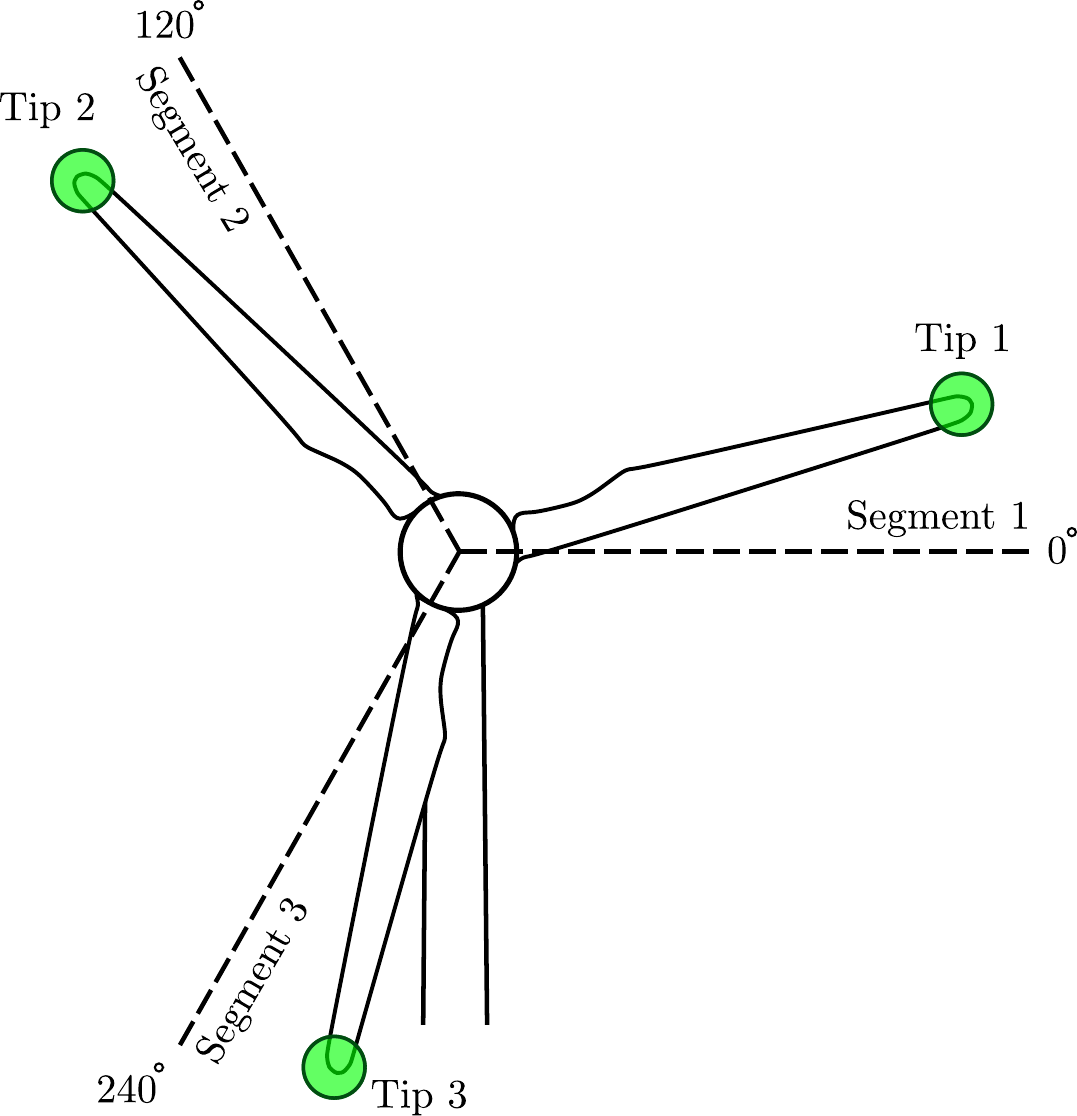}
  \caption{Wind turbine with numbered tips and numbered segments.}
  \label{fig:2_WEA_Front}
  \vspace{20pt}
\end{wrapfigure}
For the task of key point detection, we use the pretrained keypoint detection variant of YOLOv11~\cite{jocher_ultralytics_2024}. This model serves as an ideal foundation for our use case due to its proven performance, scalability, and adaptability to complex visual tasks.

One of the challenges during the training involved the labeling of blade tips. Since keypoints on an image must have unique indices in the context of keypoint detection, each of the three blade tips had to be assigned a fixed identifier (tip 1, tip 2, and tip 3 as earlier shown in Fig~\ref{fig:01_WEA}), based on its angular position. Specifically, the tip in $[\SI{0}{\degree}, \SI{120}{\degree})$ segment was labeled as tip 1, the tip in $[\SI{120}{\degree}, \SI{240}{\degree})$ segment was labeled as tip 2, and the tip in $[\SI{240}{\degree}, \SI{360}{\degree})$ segment was labeled as tip 3. This is depicted in Fig.~\ref{fig:2_WEA_Front}. While this approach solves the keypoint labeling issue, it has several drawbacks. Specifically, if the WT nacelle rotates 180 degrees, the segment labels for the tips must be reassigned, otherwise, label alignment fails, and the model cannot correctly distinguish the tips. Additionally, the hard segment boundaries cause significant loss penalties for uncertain predictions near segment edges.  

To address this issue, we modified the loss function for the three blade tips (key points 1, 2, and 3). YOLOv11 uses the object keypoint similarity (OKS) as an evaluation metric, which was proposed by the Common Objects in Context (COCO) consortium.
During training, YOLOv11 uses an adapted variant of this OKS as a loss function. Below, we present our modification to the loss function of the blade tips. The loss function of the other key points is not modified.

Let $\hat{\mathbf{p}}=\begin{pmatrix}\hat{\mathbf{p}}_1& \hat{\mathbf{p}}_2& \hat{\mathbf{p}}_3\end{pmatrix}^\top \in \mathbb{R}^{ 3 \times 2 }$ be the prediction of the three tips of the wind turbine and $\mathbf{p}=\begin{pmatrix}\mathbf{p}_1& \mathbf{p}_2& \mathbf{p}_3\end{pmatrix}^\top \in \mathbb{R}^{3 \times 2}$ the corresponding ground truth coordinates. 
The Euclidean distance between the predicted and ground truth keypoints $d$ is:
\begin{align}
    d_i=\left\|\hat{\mathbf{p}}_i - \mathbf{p}_i\right\|_2.
\end{align}
The original OKS-based loss function is computed as the sum of the squared Euclidean distances $d_i$. However, since the exact assignment of the three key points is not important, we look for the permutation of keypoint assignments that minimizes the sum of the squared Euclidean distances $d_i$ for $i=\{1,2,3\}$. There are $3!=6$ possible permutations:
\begin{align*}
\pi_1 = \begin{pmatrix}
1 & 2 & 3 \\
1 & 2 & 3
\end{pmatrix},
\pi_2 = \begin{pmatrix}
1 & 2 & 3 \\
2 & 1 & 3
\end{pmatrix},
\pi_3 = \begin{pmatrix}
1 & 2 & 3 \\
3 & 2 & 1
\end{pmatrix},
\pi_4 = \begin{pmatrix}
1 & 2 & 3 \\
1 & 3 & 2
\end{pmatrix},
\pi_5 = \begin{pmatrix}
1 & 2 & 3 \\
2 & 3 & 1
\end{pmatrix},
\pi_6 = \begin{pmatrix}
1 & 2 & 3 \\
3 & 1 & 2
\end{pmatrix}.
\end{align*}
The optimal permutation of the key points $\pi^*$ with the minimum Euclidean distance between prediction and ground truth is:
\begin{align}
    \pi^* = \argmin_{\pi_1,\ldots,\pi_6}  \begin{pmatrix}1&1&1\end{pmatrix}\left( P_{\pi_i}\hat{\mathbf{p}}-\mathbf{p}\right)^\top\left( P_{\pi_i}\hat{\mathbf{p}}-\mathbf{p}\right),
\end{align}
where $P_{\pi_i}$ is a permutation matrix corresponding to a permutation ${\pi_i}$. The OKS loss is then calculated using the optimal permuted assignment $\hat{\mathbf{p}}^* = P_{\pi^*}\hat{\mathbf{p}}$.  This modification makes the loss function invariant to the fixed ordering of the blade tips and allows the network to learn their visual localization independently of their angular position. The evaluation metric for the key points has also been adjusted accordingly.

\section{RESULTS AND EVALUATION}\label{results}
\subsection{Validation Setup}
Data generation and model training were conducted on an Ubuntu workstation equipped with an AMD Ryzen 9 3900X 12-core processor, 32 GB of DDR4 RAM, and an NVIDIA GeForce RTX 3080 GPU with 10 GB of GDDR6X memory, running NVIDIA driver version 570.144 and CUDA 12.8. A total of 12977 training and 3273 validation images were generated using the toolbox, with rendering taking approximately 270 minutes. Two YOLOv11 models, size medium (m) and small (s), were trained to assess accuracy, inference speed, and generalization performance on different network sizes. YOLOv11 m and s have 20.1 M and 9.4 M parameters, respectively. The models were each trained for 150 epochs, with the training for the model m taking 8h 4min and for model s 4h 22min. 

For the validation, two datasets are used. The first dataset consists of 3273 images generated by the toolbox. The second dataset has 83 randomly selected real-world WT images, downloaded from \cite{publicdomainpictures_wind_nodate} under the CC0 1.0 Universal Public Domain Dedication, from \cite{kaggle_object_nodate} under the Creative Commons Attribution 4.0 International (CC BY 4.0) license, from \cite{unsplash_wind_nodate} under Unsplash license, and from \cite{adobe_adobe_nodate} under Adobe Stock Standard License. These images were cropped and annotated manually. It is important to mention that the WT models used in the second dataset were not used during training and share no deliberate similarities with those seen by the network during the training. This is a strategy to evaluate the network's generalization ability on previously unseen WT types. Tab. 1 summarizes the parameters introduced in Sec.~\ref{Data generation} and lists their values that were used to generate images for this work.

\begin{table}[ht]
\centering
\caption{Experiment Parameter}
\label{tab:Parameter}
\begin{tabular}{lcl}
\toprule
\textbf{Parameter} & \textbf{Symbol} & \textbf{Sample/Distribution} \\
\midrule
Solar azimuth & $\phi_\text{s}$ & $\SI{0}{\degree}$ \\
Solar altitude & $\theta_\text{s}$ & $\SI{90}{\degree}$ \\
Dust density & $d_\text{d}$ & $1.0$ \\
Number of WTs & $n$ & $\sim U(\{1,1,1,1,1,1,2,2,2,3,3,4\})$ \\
Position of each WT & $y$ & $\sim U(\SI{0}{\meter},\SI{200}{\meter})$ or $ \sim U(\SI{0}{\meter},\SI{800}{\meter})$ equally distributed\\
 & $x$ & $\sim U(\SI{20}{\meter},y)$ or $\sim U(\SI{-20}{\meter},-y)$ equally distributed \\
Yaw rotation & $\psi$ & $\sim \mathcal{N}(\psi_\mu,\SI{5}{\degree})$ where $\psi_\mu\sim U(0,\SI{360}{\degree})$ \\
Blade rotation & $\phi$ & $\sim U(\SI{0}{\degree},\SI{360}{\degree})$ \\
Camera distance & $d_\text{c}$ & $\sim U(\SI{80}{\meter},\SI{200}{\meter})$ or $ \sim U(\SI{80}{\meter},\SI{800}{\meter})$ equally distributed \\
Camera height & $h_\text{c}$ & $\sim U(\SI{10}{\meter},\SI{260}{\meter})$ \\
Camera focal length & $f_\text{c}$ & $\sim U(\SI{3}{\milli\meter},\SI{55}{\milli\meter})$ \\
Camera roll & $\phi_\text{c}$ & $\sim\mathcal{N}(\SI{0}{\degree},\SI{3}{\degree})$ \\
Camera pitch & $\theta_\text{c}$ & calculated to vertically align the point $(\SI{0}{\meter},\SI{0}{\meter},\SI{89}{\meter})$ \\
Camera yaw & $\psi_\text{c}$ & $\SI{0}{\degree}$ \\
HSV shift foreground/background & $H_\text{f/b}$ & $\sim\mathcal{N}(0,10)$\\
 & $S_\text{f/b}$ & $\sim\mathcal{N}(0,10)$ \\
 & $V_\text{f/b}$ & $1+v_\text{f/b}$, where $v_\text{f/b}\sim\mathcal{N}(0,0.3)$\\
JPEG compression factor & $c_\text{JPEG}$ & $\sim U(45,100)$ executed on 40\% of the images \\
Noise per pixel & $\mu_\text{n}$, $\sigma_\text{n}$ & $0,\sim U(1,8)$ executed on 40 \% of the images \\
Random noise background & $n_\text{r}$ & $0.1$, when the RGB values of the noise background are \\
& &  $R\sim U(0,255),G\sim U(0,255),B\sim U(0,255).$\\
\bottomrule
\end{tabular}
\end{table}

\subsection{Validation Results and Discussion}
\begin{figure}[b]
   \centering
    \includegraphics[width=\textwidth]{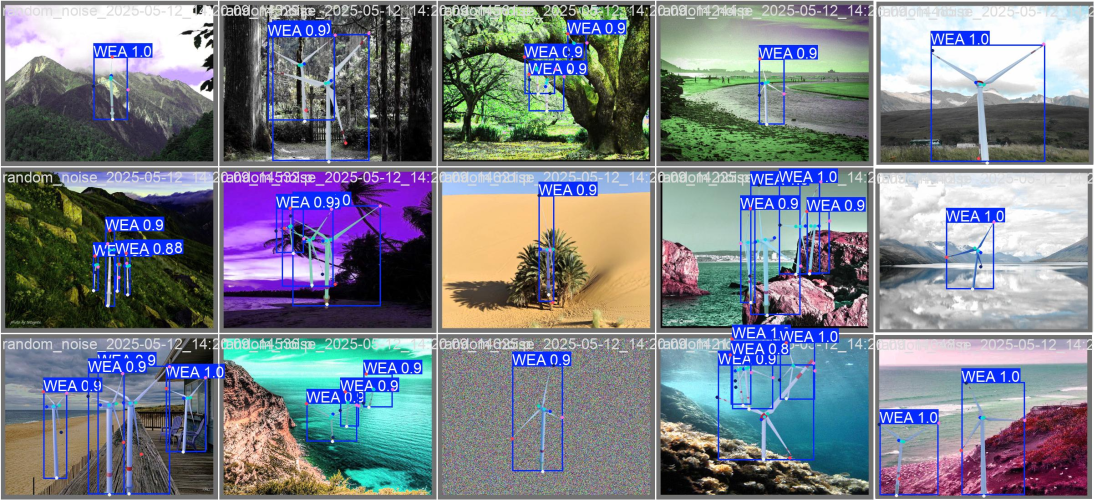}
	\caption{Example detection results from the first validation set using the trained YOLOv11 m model. Bounding boxes are drawn with class labels and confidence scores. Background images from~\cite{kaggle_landscape_nodate} under CC0 Public Domain license.}
\label{fig:6_Validationset1} 
\end{figure}

\begin{figure}[t]
   \centering
    \includegraphics[width=13cm]{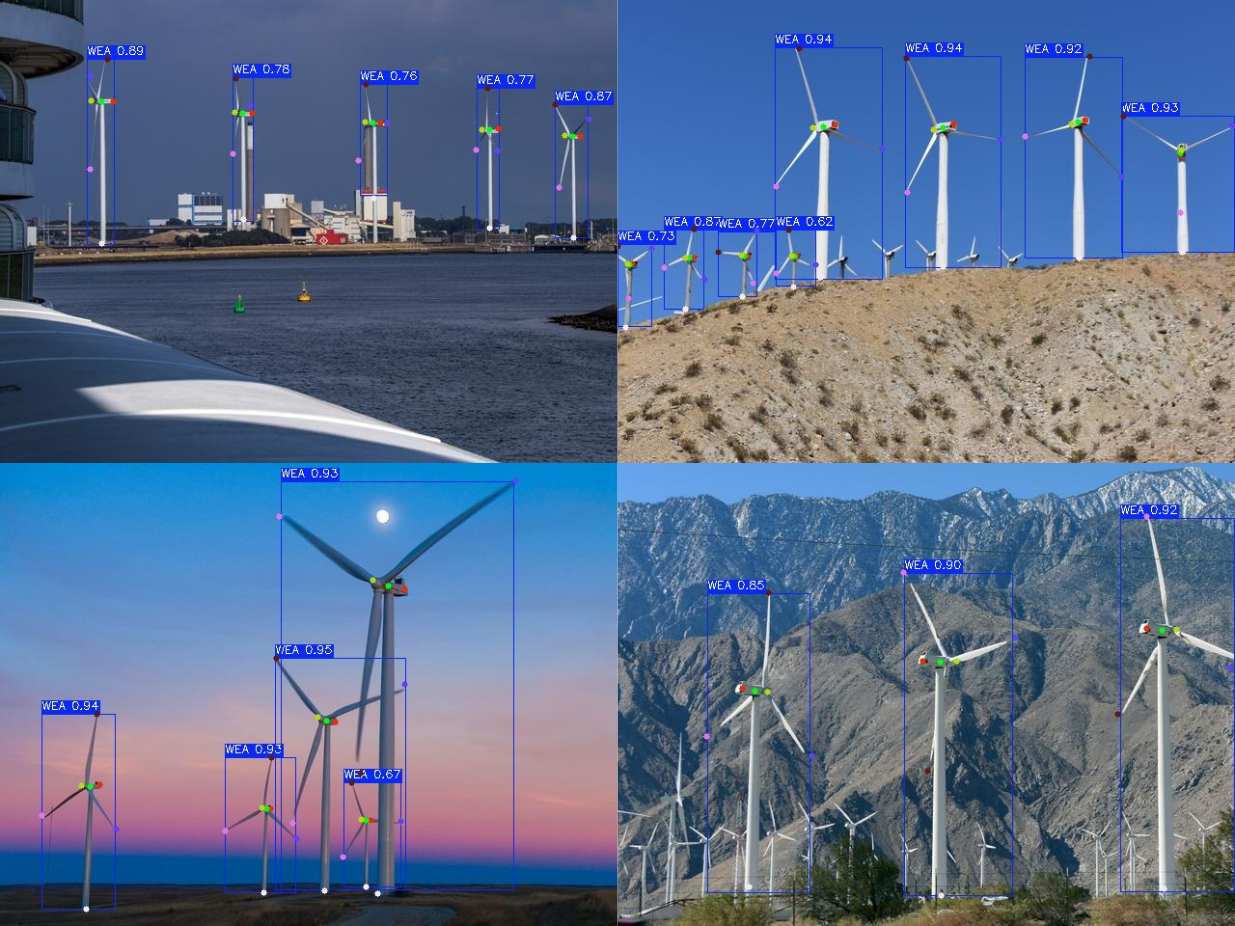}
	\caption{Example detection results from the second (real-world) validation set using the trained YOLOv11 m model. Raw images from~\cite{unsplash_wind_nodate} under Unsplash license (image lower left) and from~\cite{publicdomainpictures_wind_nodate} under CC0 Public Domain license (the rest of the images).}
\label{fig:6_Validationset2} 
\end{figure}

Fig. \ref{fig:6_Validationset1} and \ref{fig:6_Validationset2} illustrate the example inference results using both the synthetic and the real-world validation datasets, respectively. Tab. 2 shows the results on the synthetic and real-world validation sets for models m and s in terms of Mean Average Precision (mAP) and Intersection over Union (IoU), and Tab. 3 lists the inference and total latency time of the models. 

The results indicate acceptable results for both models on synthetic and real-world datasets. Noticeably, there is a drop in the mAP50-95 Box compared to other mAP scores, especially in the real-world results. This is expected, particularly in bounding box detection, where even small deviations in the size or alignment of the box are penalized at high IoU levels. In contrast, pose estimation results remain consistently high even under higher thresholds, which suggests a robustness in keypoint detection. 

Comparing the synthetic and the real-world validations indicates the compatibility between the synthetic and real-world results and strong generalization of the approach, although the WT types in the real-world data are not included in the synthetic training data, and the models were trained solely on synthetic images. However, the scores of the real-world validation are slightly better than the synthetic validation in some cases. In our opinion, this is due to the limited number of the real-world dataset, as it contains only 83 images. While these images were selected carefully from four different sources to ensure diversity, due to the limited size, the dataset cannot reach the same complexity spectrum and diversity as the synthetic dataset. As a result, we expect that expanding the real-world dataset would result in lower and more realistic performance scores.

These results also indicate that models s and m achieve comparable accuracies, with model s reaching up to \%0.3 better scores in the real-world dataset. This can be explained by the limited size and complexity of the real-world validation dataset, where both models likely reached their performance limits. However, we expect model m to perform better in larger, more diverse scenarios. The inference and total latency time of the model s are significantly lower, making this model the preferable choice for on-board deployment where computational efficiency is crucial due to the limited electrical power.

One potential limitation is that the current validation sets consist solely of still images and not video frames. In the real-world scenario of this project, video-based inputs are used that could have different characteristics, such as motion blur due to factors like shutter speed. 
At the time of evaluation, video validation was not possible, as we did not have access to labeled video data, and manual annotation was very time-consuming.

\begin{table}[ht]\label{tab:onlyone}
\centering
\caption{Performance metrics on validation datasets for YOLOv11 models (s and m).}
\begin{tabular}{llcccc}
\toprule
\textbf{Model} & \textbf{Dataset} & \textbf{mAP50 Box} & \textbf{mAP50-95 Box} & \textbf{mAP50 Pose} & \textbf{mAP50-95 Pose} \\
\midrule
\multirow{2}{*}{YOLOv11-s} & Synthetic   & 0.9897 & 0.9252 & 0.9881 & 0.9731 \\
                           & Real-world  & 0.9929 & 0.8724 & 0.9929 & 0.9722 \\
\midrule
\multirow{2}{*}{YOLOv11-m} & Synthetic   & 0.9906 & 0.9419 & 0.9897 & 0.9782 \\
                           & Real-world  & 0.9910 & 0.8696 & 0.9920 & 0.9703 \\
\bottomrule
\end{tabular}
\end{table}

\begin{table}[ht!]\label{tab:inferencetime}
\centering
\caption{Inference and total latency for YOLOv11-s and YOLOv11-m .}
\begin{tabular}{l@{\hskip 8pt}cc@{\hskip 8pt}cc}
\toprule
 \multicolumn{2}{c}{\textbf{Inference time (\si{\milli\second})}} & \multicolumn{2}{c}{\textbf{Total latency (\si{\milli\second})}} \\
 \textbf{s} & \textbf{m} & \textbf{s} & \textbf{m} \\
\midrule
 2.0 & 4.6 & 2.4 & 5.1 \\
\bottomrule
\end{tabular}
\end{table}

\section{CONCLUSION AND OUTLOOK}\label{conclusion}
In this study, we presented a toolbox to generate synthetic images of wind turbines and used them exclusively to train YOLOv11 models for wind turbine keypoint detection. These findings validate the performance of the proposed data generation toolbox and highlight its potential for applications where labeled real-world data is limited or unavailable. The results further indicate that synthetic images, when carefully modeled, can replace real-world images for training deep learning models in wind turbine keypoint detection tasks. This suggests that time-consuming and expensive data collection efforts may be reduced in these scenarios.

In our future research, we aim to expand the toolbox to extract numerical data from the predictions, such as yaw angle and blade rotation speed. Furthermore, the models will be tested with larger real-world datasets and with videos.

\bibliographystyle{spiebib}
\bibliography{references}

\end{document}